\documentclass[runningheads]{llncs}

\usepackage[numbers]{natbib} 

\usepackage{graphicx, subcaption} \graphicspath{ {Images/} }

\usepackage[pdftex, pdfpagelabels, hypertexnames, bookmarks, bookmarksnumbered, plainpages=false, naturalnames=false, pdfpagemode=UseOutlines]{hyperref}
\hypersetup{colorlinks, citecolor=[rgb]{0.0,0.0,0.0}, filecolor=[rgb]{0.0,0.0,0.0}, linkcolor=[rgb]{0.0,0.0,0.0}, urlcolor=[rgb]{0.0,0.0,0.0}}

\usepackage{amsmath}
\usepackage{booktabs}
\usepackage{microtype}

\begin{document}

\title{End-to-End Boundary Aware Networks for\\ Medical Image Segmentation}

\author{Ali Hatamizadeh\inst{1,2} \and Demetri  Terzopoulos\inst{1}  \and Andriy Myronenko\inst{2}}

\authorrunning{A.~Hatamizadeh, D.~Terzopoulos, and A.~Myronenko}

\institute{Computer Science Department, University of California, Los Angeles, CA, USA
\and
NVIDIA, Santa Clara, CA, USA\\
}

\maketitle

\begin{abstract}
Fully convolutional neural networks (CNNs) have proven to be effective at representing and classifying textural information, thus transforming image intensity into output class masks that achieve semantic image segmentation. In medical image analysis, however, expert manual segmentation often relies on the boundaries of anatomical structures of interest. We propose boundary aware CNNs for medical image segmentation. Our networks are designed to account for organ boundary information, both by providing a special network edge branch and edge-aware loss terms, and they are trainable end-to-end. We validate their effectiveness on the task of brain tumor segmentation using the BraTS 2018 dataset. Our experiments reveal that our approach yields more accurate segmentation results, which makes it promising for more extensive application to medical image segmentation.   
\end{abstract}

\keywords{Medical Image Segmentation \and Semantic Segmentation \and Convolutional Neural Networks
\and Deep Learning}

\section{Introduction}
\label{sec:intro}

Deep learning approaches to semantic image segmentation have achieved state-of-the-art performance in medical image analysis~\citep{Ronneberger15,Myronenko18,hatamizadeh2019deep,hatamizadeh2019deep2}. With the advent of convolutional neural networks (CNNs), the earliest segmentation methods attempted to classify every pixel based on a corresponding image patch, which often resulted in slow inference times. Fully convolutional neural networks~\citep{Ronneberger15}, can segment the whole image at once, but the underlying assumption remained---instead of a patch, the corresponding image region (receptive field) centered on the pixel is used for the final pixel segmentation. Since convolutions are spatially invariant, segmentation networks can operate on any image size and infer dense pixel-wise segmentation. 

\citet{geirhos2018} empirically demonstrated that, unlike the human visual system, common CNN architectures are biased towards recognizing image textures, not object shape representations. In medical image analysis, however, expert manual segmentation usually relies on boundary and organ shape identification. For instance, a radiologist segmenting a liver from CT images would usually trace liver edges first, from which the internal segmentation mask is easily deduced. This observation motivates us to devise segmentation networks that prioritize the representation of edge information in anatomical structures by leveraging an additional edge module whose training is supervised by edge-aware loss functions.

Recently, several authors have pursued deep learning approaches for object edge prediction. \citet{ZhidingYu17} proposed a multilabel semantic boundary detection network to improve a wide variety of vision tasks by predicting edges directly, including a new skip-layer architecture in which category-wise edge activations at the top convolution layer share and are fused with the same set of bottom layer features, along with a multilabel loss function to supervise the fused activations. Subsequently, \citet{yu2018seal} showed that label misalignment can cause considerably degraded edge learning quality, and addressed this issue by proposing a simultaneous edge alignment and learning framework. \citet{Acuna_2019_CVPR} predicted object edges by identifying pixels that belong to class boundaries, proposing a new layer and a loss that enforces the detector to predict a maximum response along the normal direction at an edge, while also regularizing its direction. \citet{takikawa2019gated} proposed gated-shape CNNs for semantic segmentation of natural images in which such gates are employed to remove the noise from higher-level activations and process the relevant boundary-related information separately. Aiming to learn semantic boundaries, \citet{HuPanoptic2019} presented a framework that aggregates different tasks of object detection, semantic segmentation,
and instance edge detection into a single holistic network with multiple branches, demonstrating significant improvements over conventional approaches through end-to-end training.

In the present paper, we introduce an encoder-decoder architecture that leverages a special interconnected edge layer module that is supervised by edge-aware losses in order to preserve boundary information and emphasize it during training. By explicitly accounting for the edges, we encourage the network to internalize edge importance during training. Our method utilizes edge information only to assist training for semantic segmentation, not for the main purpose of predicting edges directly. This strategy enables a structured regularization mechanism for our network during training and results in more accurate and robust segmentation performance during inference. We validate the effectiveness of our network on the task of brain tumor segmentation using the BraTS 2018 dataset~\citep{brats2}.

\section{Methods}
\label{sec:methods}

\subsection{Architecture}

\begin{figure}[t]
\includegraphics[width=\textwidth]{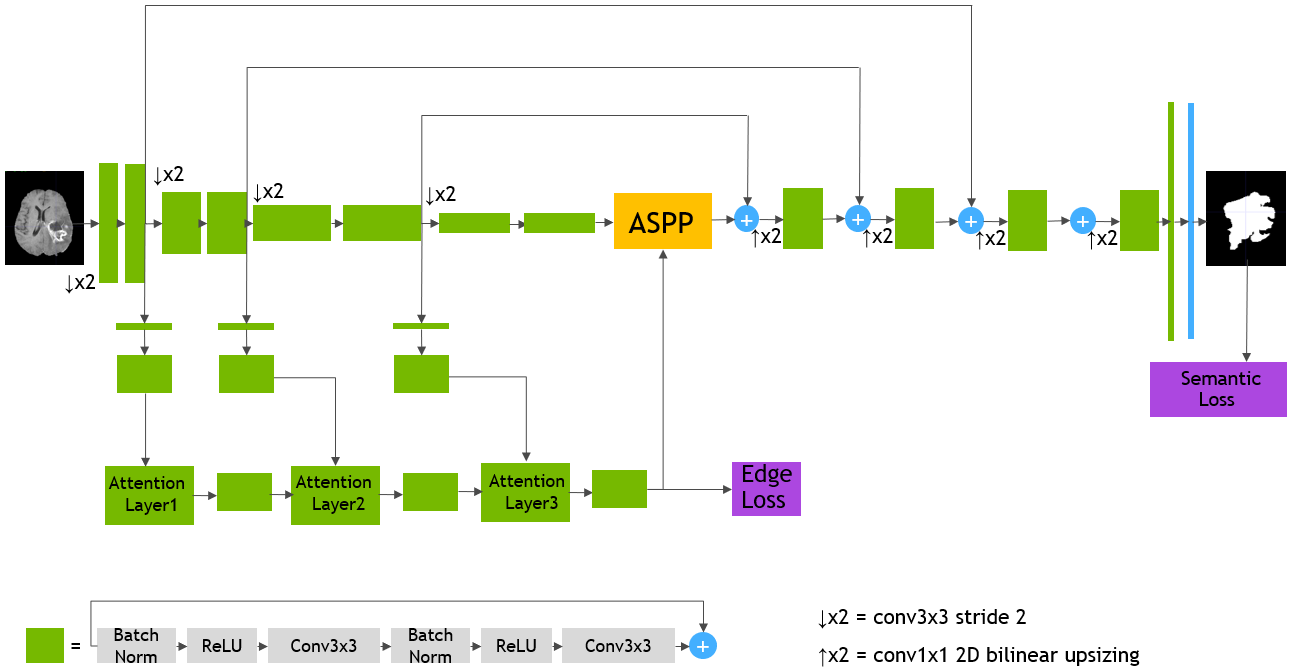}
  \caption{Our 2D fully convolutional architecture. We use dilated spatial pyramid pooling to effectively aggregate the outputs of different stages.}
  \label{fig:pipeline}
\end{figure}

Our network comprises a main encoder-decoder stream for semantic segmentation as well as a shape stream that processes the feature maps at the boundary level (Fig.~\ref{fig:pipeline}). In the encoder portion of the main stream, every resolution level includes two residual blocks whose outputs are fed to the corresponding resolution of the shape stream. A $1\times1$ convolution is applied to each input to the shape stream and the result is fed into an attention layer that is discussed in the next section. The outputs of the first two attention layers are fed into connection residual blocks. The output of the last attention layer is concatenated with the output of the encoder in the main stream and fed into a dilated spatial pyramid pooling layer. Losses that contribute to tuning the weights of the model come from the output of the shape stream that is resized to the original image size, as well as the output of the main stream. 

\subsection{Attention Layer}
Each attention layer receives inputs from the previous attention layer as well as the main stream at the corresponding resolution. Let $s_l$ and $m_l$ denote the attention layer and main stream layer inputs at resolution $l$. We first concatenate $s_l$ and $m_l$ and apply a $1\times1$ convolution layer $C_{1 \times 1}$ followed by a sigmoid function $\sigma$ to obtain an attention map
\begin{equation}
\alpha_{l}=\sigma\big(C_{1 \times 1}(s_{l}\mathbin\Vert m_{l})\big).
\label{eq:att1}
\end{equation} 
An element-wise multiplication is then performed with the input to the attention layer to obtain the output of the attention layer, denoted as
\begin{equation}
o_{l}= s_{l} \odot \alpha_{l}.
\label{eq:att2}
\end{equation} 

\subsection{Boundary Aware Segmentation }

Our network jointly learns the semantics and boundaries by supervising the output of the main stream as well as the edge stream. We use the generalized Dice loss on predicted outputs of the main stream and the shape stream. Additionally, we add a weighted binary cross entropy loss to the shape stream loss in order to deal with the large imbalance between the boundary and non-boundary pixels. The overall loss function of our network is
\begin{equation}
L_\textrm{total}=\lambda_{1} L_\textrm{Dice}(y_\textrm{pred},y_\textrm{true})+ \lambda_{2}L_\textrm{Dice}(s_\textrm{pred},s_\textrm{true})+ \lambda_{3} L_\textrm{Edge}(s_\textrm{pred},s_\textrm{true}), 
\label{eq:finalloss}
\end{equation}
where $y_\textrm{pred}$ and $y_\textrm{true}$ denote the pixel-wise semantic predictions of the main stream while $s_\textrm{pred}$ and $s_\textrm{true}$ denote the boundary predictions of the shape stream; $s_\textrm{true}$ can be obtained by computing the spatial gradient of $y_\textrm{true}$.

The Dice loss \citep{Milletari16} in (\ref{eq:finalloss}) is
\begin{equation}
L_\textrm{Dice}= 1- \frac{2 \sum y_\textrm{true} y_\textrm{pred} }{\sum y_\textrm{true}^2 + \sum y_\textrm{pred}^2 + \epsilon},
\label{eq:dice}
\end{equation} 
where summation is carried over the total number of pixels and $\epsilon$ is a small constant to prevent division by zero.

The edge loss in (\ref{eq:finalloss}) is
\begin{equation}
L_\textrm{Edge}= -\beta \sum_{j\in y_{+}} \log P(y_{\textrm{pred},j}=1|x;\theta)-(1-\beta) \sum_{j\in y_{-}} \log P(y_{\textrm{pred},j}=0|x;\theta),
\label{eq:bce}
\end{equation} 
where $x$, $\theta$, $y_{-}$, and $y_{+}$ denote the input image, CNN parameters, and edge and non-edge pixel sets, respectively, $\beta$ is the ratio of non-edge pixels over the entire number of pixels, and $P(y_{\textrm{pred},j})$ denotes the probability of the predicated class at pixel $j$. 

\section{Experiments}

\subsection{Datasets}

In our experiments, we used the BraTS 2018~\citep{brats2}, which provides multimodal 3D brain MRIs and ground truth brain tumor segmentations annotated by physicians, consisting of 4 MRI modalities per case (T1, T1c, T2, and FLAIR). Annotations include 3 tumor subregions---the enhancing tumor, the peritumoral edema, and the necrotic and non-enhancing tumor core. The annotations were combined into 3 nested subregions---whole tumor (WT), tumor core (TC), and enhancing tumor (ET). The data were collected from 19 institutions, using various MRI scanners. For simplicity, we use only a single input MRI modality (T1c) and aim to segment a single tumor region---TC, which includes the main tumor components (nectrotic core, enhancing, and non-enhancing tumor regions). Furthermore, even though the original data is 3D ($240\times240\times155$), we operate on 2D slices for simplicity. We have extracted several axial slices centered around the tumor region from each 3D volume, and combined them into a new 2D dataset.

\subsection{Implementation Details}

We have implemented our model in Tensorflow. The brain input images were resized to predefined sizes of $240\times240$ and normalized to the intensity range $[0,1]$. The model was trained on NVIDIA Titan RTX and an Intel® Core™ i7-7800X CPU @ 3.50GHz $\times$ 12 with a batch size of 30 for all models. We used $\lambda_{1}=1.0$, $\lambda_{2}=0.5$, and $\lambda_{3}=0.1$ in (\ref{eq:finalloss}). The Adam optimization algorithm was used with initial learning rate of $ \alpha_{0} = 1.0^{-3}$ and further decreased according to
\begin{equation}
\label{eq:learningrate}
\alpha = \alpha_{0}\left(1-e/N_{e}\right)^{0.9}, 
\end{equation}
where $e$ denotes the current epoch and $N_{e}$ the total number of epochs, following~\citep{Myronenko18}. We have evaluated the performance of our model by using the Dice score, Jaccard index, and Hausdorff distance.


\section{Results and Discussion}
\label{sec:results}

\begin{figure}[t]
\def\x{0.24}

\includegraphics[width=\x\linewidth,height=\x\linewidth]{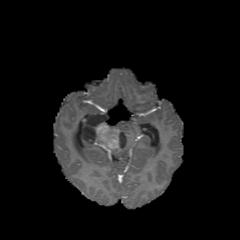}
\hfill
\includegraphics[width=\x\linewidth,height=\x\linewidth]{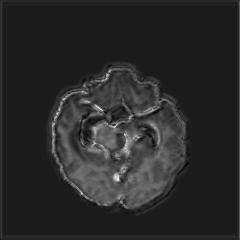}
\hfill
\includegraphics[width=\x\linewidth,height=\x\linewidth]{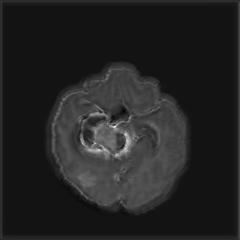}
\hfill
\includegraphics[width=\x\linewidth,height=\x\linewidth]{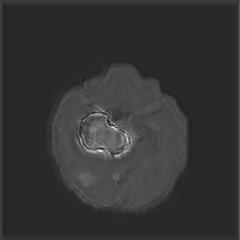}

\vspace{0.75pt}

\includegraphics[width=\x\linewidth,height=\x\linewidth]{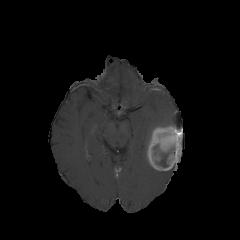}
\hfill
\includegraphics[width=\x\linewidth,height=\x\linewidth]{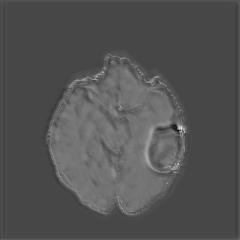}
\hfill
\includegraphics[width=\x\linewidth,height=\x\linewidth]{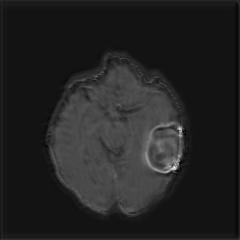}
\hfill
\includegraphics[width=\x\linewidth,height=\x\linewidth]{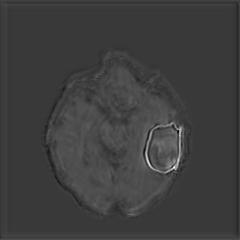}

\vspace{0.75pt}

\includegraphics[width=\x\linewidth,height=\x\linewidth]{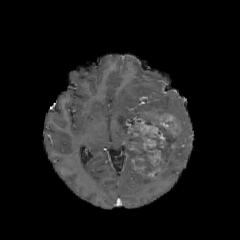}
\hfill
\includegraphics[width=\x\linewidth,height=\x\linewidth]{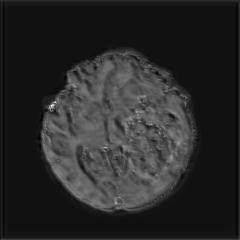}
\hfill
\includegraphics[width=\x\linewidth,height=\x\linewidth]{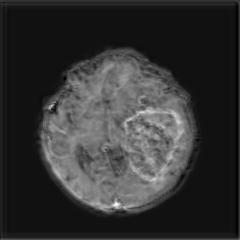}
\hfill
\includegraphics[width=\x\linewidth,height=\x\linewidth]{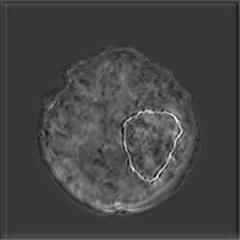}

\makebox[\x\linewidth]{(a)} \hfill \makebox[\x\linewidth]{(b)} \hfill
\makebox[\x\linewidth]{(c)} \hfill \makebox[\x\linewidth]{(d)}

\caption{(a) Input image. Outputs of : (b) Attention Layer~1. (c) Attention Layer~2. (d) Attention Layer~3.The boundary emphasis becomes more prominent in the subsequent attention layers.}
\label{attention_maps}
\end{figure}

\paragraph{Boundary Stream:}
Fig.~\ref{attention_maps} demonstrates the output of each of the attention layers in our dedicated boundary stream. In essence, each attention layer progressively localizes the tumor and refines the boundaries. The first attention layer has learned rough estimate of the boundaries around the tumor and localized it, whereas the second and third layers have learned more fine-grained details of the edges and boundaries, refining the localization. Moreover, since our architecture leverages a dilated spatial pyramid pooling to merge the learned feature maps of the regular segmentation stream and the boundary stream, multiscale regional and boundary information have been preserved and fused properly, which has enabled our network to capture the small structural details of the tumor. 

\begin{table}[t]
\centering
\begin{tabular}{llll}
\toprule
Model & Dice Score & Jaccard Index &  Hausdorff Distance \\
\midrule
U-Net~\citep{Ronneberger15} & 0.731$\pm$0.230   & 0.805~$\pm$0.130  & 3.861$\pm$1.342     \\
V-Net~\citep{Milletari16} & 0.769$\pm$0.270  &0.837$\pm$0.140  & 3.667$\pm$1.329 \\
Ours (no edge loss)~~ & 0.768$\pm$0.236   & 0.832$\pm$0.136  &  3.443$\pm$1.218    \\
Ours  & \textbf{0.822$\pm$0.176~~} & \textbf{0.861$\pm$0.112~~} & \textbf{3.406$\pm$1.196} \\
\bottomrule
\end{tabular}
\medskip
\caption{Performance evaluations of different models. We validate the contribution of the edge loss by measuring performance with and without this layer}
\label{res}
\end{table}

\paragraph{Edge-Aware Losses:}
To validate the effectiveness of the loss supervision, we have trained our network without enforcing the supervision of the edge loss during the learning process, but with the same architecture. Table~\ref{res} shows that our network performs very similarly to V-Net~\citep{Milletari16} without edge supervision, since ours employs similar residual blocks as V-Net in its main encoder-decoder, and its boundary stream does not seem to contribute to the learning of useful features for segmentation.
In essence, the boundary stream also impacts the down-stream layers of the encoder by emphasizing edges during training.

\begin{figure}
\def\x{0.19}

\includegraphics[width=\x\linewidth,height=\x\linewidth]{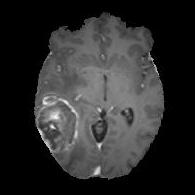}
\hfill
\includegraphics[width=\x\linewidth,height=\x\linewidth]{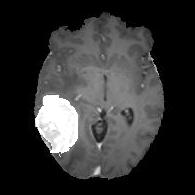}
\hfill
\includegraphics[width=\x\linewidth,height=\x\linewidth]{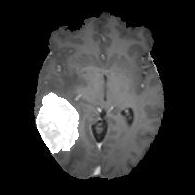}
\hfill
\includegraphics[width=\x\linewidth,height=\x\linewidth]{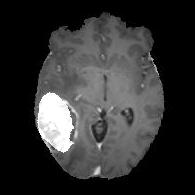}
\hfill
\includegraphics[width=\x\linewidth,height=\x\linewidth]{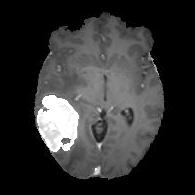}

\vspace{0.75pt}

\includegraphics[width=\x\linewidth,height=\x\linewidth]{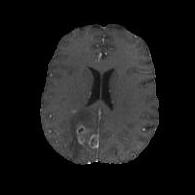}
\hfill
\includegraphics[width=\x\linewidth,height=\x\linewidth]{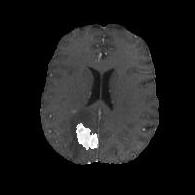}
\hfill
\includegraphics[width=\x\linewidth,height=\x\linewidth]{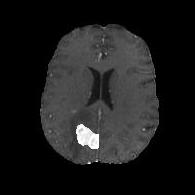}
\hfill
\includegraphics[width=\x\linewidth,height=\x\linewidth]{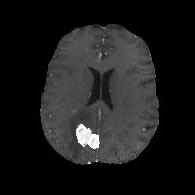}
\hfill
\includegraphics[width=\x\linewidth,height=\x\linewidth]{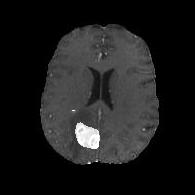}

\vspace{0.75pt}

\includegraphics[width=\x\linewidth,height=\x\linewidth]{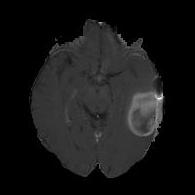}
\hfill
\includegraphics[width=\x\linewidth,height=\x\linewidth]{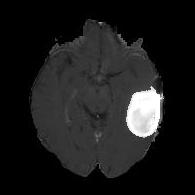}
\hfill
\includegraphics[width=\x\linewidth,height=\x\linewidth]{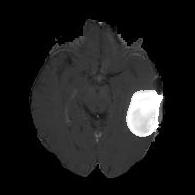}
\hfill
\includegraphics[width=\x\linewidth,height=\x\linewidth]{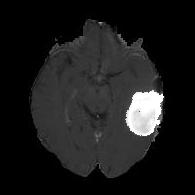}
\hfill
\includegraphics[width=\x\linewidth,height=\x\linewidth]{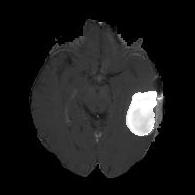}

\vspace{0.75pt}

\includegraphics[width=\x\linewidth,height=\x\linewidth]{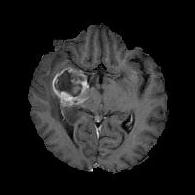}
\hfill
\includegraphics[width=\x\linewidth,height=\x\linewidth]{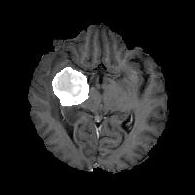}
\hfill
\includegraphics[width=\x\linewidth,height=\x\linewidth]{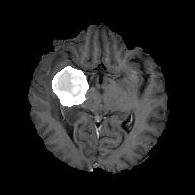}
\hfill
\includegraphics[width=\x\linewidth,height=\x\linewidth]{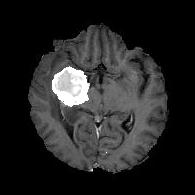}
\hfill
\includegraphics[width=\x\linewidth,height=\x\linewidth]{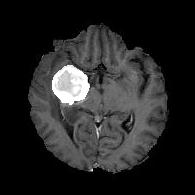}

\vspace{0.75pt}

\includegraphics[width=\x\linewidth,height=\x\linewidth]{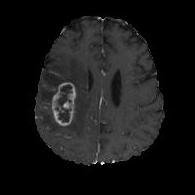}
\hfill
\includegraphics[width=\x\linewidth,height=\x\linewidth]{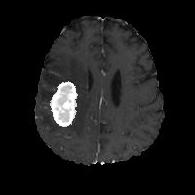}
\hfill
\includegraphics[width=\x\linewidth,height=\x\linewidth]{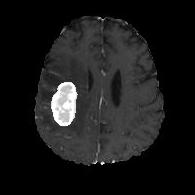}
\hfill
\includegraphics[width=\x\linewidth,height=\x\linewidth]{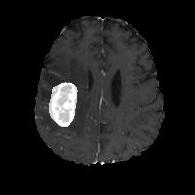}
\hfill
\includegraphics[width=\x\linewidth,height=\x\linewidth]{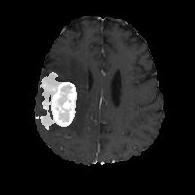}

\vspace{0.75pt}

\includegraphics[width=\x\linewidth,height=\x\linewidth]{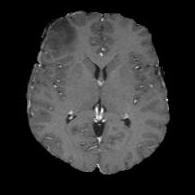}
\hfill
\includegraphics[width=\x\linewidth,height=\x\linewidth]{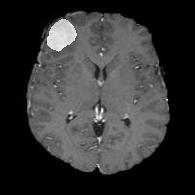}
\hfill
\includegraphics[width=\x\linewidth,height=\x\linewidth]{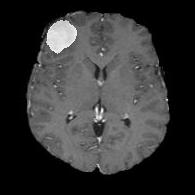}
\hfill
\includegraphics[width=\x\linewidth,height=\x\linewidth]{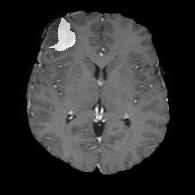}
\hfill
\includegraphics[width=\x\linewidth,height=\x\linewidth]{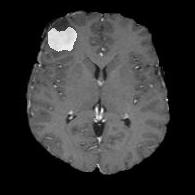}

\vspace{0.75pt}

\includegraphics[width=\x\linewidth,height=\x\linewidth]{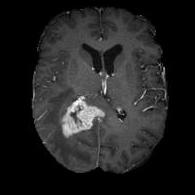}
\hfill
\includegraphics[width=\x\linewidth,height=\x\linewidth]{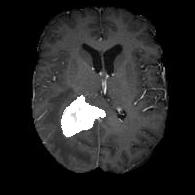}
\hfill
\includegraphics[width=\x\linewidth,height=\x\linewidth]{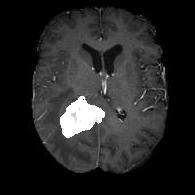}
\hfill
\includegraphics[width=\x\linewidth,height=\x\linewidth]{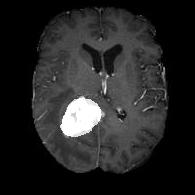}
\hfill
\includegraphics[width=\x\linewidth,height=\x\linewidth]{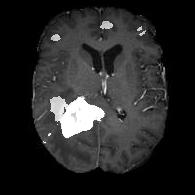}

\vspace{0.75pt}

\makebox[\x\linewidth]{(a)} \hfill \makebox[\x\linewidth]{(b)} \hfill
\makebox[\x\linewidth]{(c)} \hfill \makebox[\x\linewidth]{(d)} \hfill \makebox[\x\linewidth]{(e)}

\caption{(a) Input images. (b) Labels. (c) Ours. (d) V-Net. (e) U-Net.}
\label{fig_data}
\end{figure}

\paragraph{Comparison to Competing Methods:}
We have compared the performance of our model against the most popular deep learning-based semantic segmentation networks, U-Net~\citep{Ronneberger15} and V-Net~\citep{Milletari16} (Fig.~\ref{fig_data}). Our model outperforms both by a considerable margin in all evaluation metrics. In particular, U-Net performs poorly in most cases due to the high false positive of its segmentation predictions, as well as the imprecision of its boundaries. The powerful residual block in the  V-Net architecture seems to alleviate these issues to some extent, but V-Net also fails to produce high-quality boundary predictions. The emphasis of learning useful edge-related information during the training of our network appears to effectively regularize the network such that boundary accuracy is improved. 

\section{Conclusion}
\label{sec:conclusion}

We have proposed an end-to-end-trainable boundary aware network for joint semantic segmentation of medical images. Our network explicitly accounts for object edge information by using a dedicated shape stream that processes the feature maps at the boundary level and fuses the multiscale contextual information of the boundaries with the encoder output of the regular segmentation stream. Additionally, edge-aware loss functions emphasize learning of the edge information during training by tuning the weights of the downstream encoder and regularizing the network to prioritize boundaries. We have validated the effectiveness of our approach on the task of brain tumor segmentation using the BraTS 2018 dataset. Our results indicate that our network produces more accurate segmentation outputs with fine-grained boundaries in comparison to the popular segmentation networks U-Net and V-Net.

\bibliography{mlmi19}

\begin{thebibliography}{12}
\providecommand{\natexlab}[1]{#1}
\providecommand{\url}[1]{\texttt{#1}}
\providecommand{\urlprefix}{}

\bibitem[{Acuna et~al.(2019)Acuna, Kar, and Fidler}]{Acuna_2019_CVPR}
Acuna, D., Kar, A., Fidler, S.: Devil is in the edges: Learning semantic
  boundaries from noisy annotations.
\newblock In: The IEEE Conference on Computer Vision and Pattern Recognition
  (CVPR) (2019)

\bibitem[{Bakas et~al.(2017)Bakas, Akbari, Sotiras, Bilello, Rozycki, Kirby,
  Freymann, Farahani, and Davatzikos}]{brats2}
Bakas, S., Akbari, H., Sotiras, A., Bilello, M., Rozycki, M., Kirby, J.,
  Freymann, J., Farahani, K., Davatzikos, C.: Advancing the cancer genome atlas
  glioma {MRI} collections with expert segmentation labels and radiomic
  features.
\newblock Scientific Data 4 (2017)

\bibitem[{Geirhos et~al.(2019)Geirhos, Rubisch, Michaelis, Bethge, Wichmann,
  and Brendel}]{geirhos2018}
Geirhos, R., Rubisch, P., Michaelis, C., Bethge, M., Wichmann, F.A., Brendel,
  W.: Imagenet-trained {CNN}s are biased towards texture; increasing shape bias
  improves accuracy and robustness.
\newblock In: International Conference on Learning Representations (ICLR)
  (2019)

\bibitem[{Hatamizadeh et~al.(2019{\natexlab{a}})Hatamizadeh, Hoogi, Sengupta,
  Lu, Wilcox, Rubin, and Terzopoulos}]{hatamizadeh2019deep2}
Hatamizadeh, A., Hoogi, A., Sengupta, D., Lu, W., Wilcox, B., Rubin, D.,
  Terzopoulos, D.: Deep active lesion segmentation.
\newblock arXiv preprint arXiv:1908.06933  (2019{\natexlab{a}})

\bibitem[{Hatamizadeh et~al.(2019{\natexlab{b}})Hatamizadeh, Hosseini, Liu,
  Schwartz, and Terzopoulos}]{hatamizadeh2019deep}
Hatamizadeh, A., Hosseini, H., Liu, Z., Schwartz, S.D., Terzopoulos, D.: Deep
  dilated convolutional nets for the automatic segmentation of retinal vessels.
\newblock arXiv preprint arXiv:1905.12120  (2019{\natexlab{b}})

\bibitem[{Hu et~al.(2019)Hu, Zou, and Feng}]{HuPanoptic2019}
Hu, Y., Zou, Y., Feng, J.: Panoptic edge detection.
\newblock https://arxiv.org/abs/1906.00590  (2019)

\bibitem[{Milletari et~al.(2016)Milletari, Navab, and Ahmadi}]{Milletari16}
Milletari, F., Navab, N., Ahmadi, S.A.: V-net: Fully convolutional neural
  networks for volumetric medical image segmentation.
\newblock In: Fourth International Conference on 3D Vision (3DV) (2016)

\bibitem[{Myronenko(2018)}]{Myronenko18}
Myronenko, A.: {3D} {MRI} brain tumor segmentation using autoencoder
  regularization.
\newblock In: {BrainLes}, Medical Image Computing and Computer Assisted
  Intervention {(MICCAI)}. pp. 311--320. LNCS, Springer (2018)

\bibitem[{Ronneberger et~al.(2015)Ronneberger, P.Fischer, and
  Brox}]{Ronneberger15}
Ronneberger, O., P.Fischer, Brox, T.: U-net: Convolutional networks for
  biomedical image segmentation.
\newblock In: Proc. MICCAI. LNCS, vol. 9351, pp. 234--241 (2015)

\bibitem[{Takikawa et~al.(2019)Takikawa, Acuna, Jampani, and
  Fidler}]{takikawa2019gated}
Takikawa, T., Acuna, D., Jampani, V., Fidler, S.: Gated-scnn: Gated shape cnns
  for semantic segmentation.
\newblock arXiv preprint arXiv:1907.05740  (2019)

\bibitem[{Yu et~al.(2017)Yu, Feng, Liu, and Ramalingam}]{ZhidingYu17}
Yu, Z., Feng, C., Liu, M., Ramalingam, S.: Casenet: Deep category-aware
  semantic edge detection.
\newblock In: CVPR (2017)

\bibitem[{Yu et~al.(2018)Yu, Liu, Zou, Feng, Ramalingam, Vijaya~Kumar, and
  Kautz}]{yu2018seal}
Yu, Z., Liu, W., Zou, Y., Feng, C., Ramalingam, S., Vijaya~Kumar, B., Kautz,
  J.: Simultaneous edge alignment and learning.
\newblock In: European Conference on Computer Vision (ECCV) (2018)

\end{thebibliography}

\end{document}